%% file: arvix_version_v2.tex
\documentclass{article} 
\usepackage{iclr2025_conference,times}
\iclrfinalcopy
\input{math_commands.tex}

\usepackage{hyperref}
\usepackage{url}

\usepackage{amsthm,amsmath,amssymb}
\usepackage{mathrsfs}
\usepackage{bbm}
\usepackage{multirow}
\usepackage{multicol}
\usepackage{booktabs}
\usepackage{utfsym}
\usepackage{bbding}
\usepackage{graphicx}
\usepackage{enumitem}

\title{MetaTool: Facilitating Large Language Models to Master Tools with Meta-task Augmentation}

\author{Xiaohan Wang, Dian Li\thanks{Corresponding author}, Yilin Zhao, Sinbadliu, Hui Wang\\
Foundation Technology Center, Tencent PCG\\
{\tt\small \{shawnbywang, goodli, yilinnzhao, sinbadliu, joltwang\}@tencent.com}
}

\begin{document}

\maketitle
\begin{abstract}
\begin{quote}
Utilizing tools with Large Language Models (LLMs) is essential for grounding AI agents in real-world applications. The prevailing approach involves few-shot prompting with demonstrations or fine-tuning with expert annotations. However, mere in-context demonstrations may fail to cover sufficient knowledge for complex tools and tasks. Training on solution paths is also hindered by the high cost of expert annotations and generalizing to new tools. A core challenge of generalizable tool use lies in understanding the ``meta'', or fundamental natures of tools that are transferable across tasks, such as causality and constraints. In this paper, we present \textit{MetaTool}, a novel tool learning methodology designed to generalize across any reusable toolset. Our approach incorporates a self-supervised augmentation technique derived from a series of meta-tasks. This involves predicting masked elements in the tool execution process. The self-supervised procedure enables scalable generation of high-quality QA data, which is handy for supervising tool understanding. By incorporating meta-task data into task-oriented training, our method significantly enhances the performance of open-source LLMs, achieving results comparable to ChatGPT in both tool-based planning and chatting scenarios. Through large-scale instruction tuning, the MetaTool model demonstrates impressive zero-shot generalizability on new tasks.
\end{quote}
\end{abstract}

\section{Introduction} 

Distinguished from other species, a unique characteristic of human advanced intelligence is using complex tools, which expands the frontiers neural intelligence can reach. With the advent of powerful foundation models, AI has the potential to solve complex tasks with these external mechanisms. LLMs have been majorly oriented towards either tool-augmented chatbots equipped with retrievers and search engines, or tool-oriented agents (e.g. web navigation \cite{rawles2023android,hong2024cogagent}, embodied manipulation \cite{chi2023diffusion}) that achieve task objectives through tool output \cite{qin2023tool}. While the former emphasizes generalizing to various tools, the latter focuses on complex tools and scenarios.

To efficiently integrate LLMs with tools, a mainstream way relies on in-context learning (ICL). The model is provided with the ``cookbook'' of tools in zero-shot prompting or demonstrations in few-shot prompting \cite{xu2023tool,paranjape2023art,brown2020language}. It may work well on simple tools with frameworks like LangChain \cite{langchain}. However, for complex tasks with sophisticated tools, in-context learning is limited that demonstrations can not exhaust all scenarios, and manuals are also limited in length. Ultimately, it's impractical to expect LLMs to be intelligent enough to master any tool without the experience of using it. 
On the other side, training-based methods \cite{qin2023toolllm,patil2023gorilla,dubey2024llama} mainly adopt supervised fine-tuning with annotated expert solutions. Despite the difficulties in scaling up the optimal annotation, supervision with task solutions has limitations. Task-agnostic knowledge of tools can be neglected, which hinders the generalization to diverse scenarios or new tools. Self-play training methods like Toolformer \cite{schick2024toolformer} and TALM \cite{parisi2022talm} integrate the inference process with self-supervised tool calling data. Although such manner maintains the generality of tool calling, it's constrained in question-answering scenarios.

\begin{figure}[tb]
    \centering
    \includegraphics[width=1\linewidth]{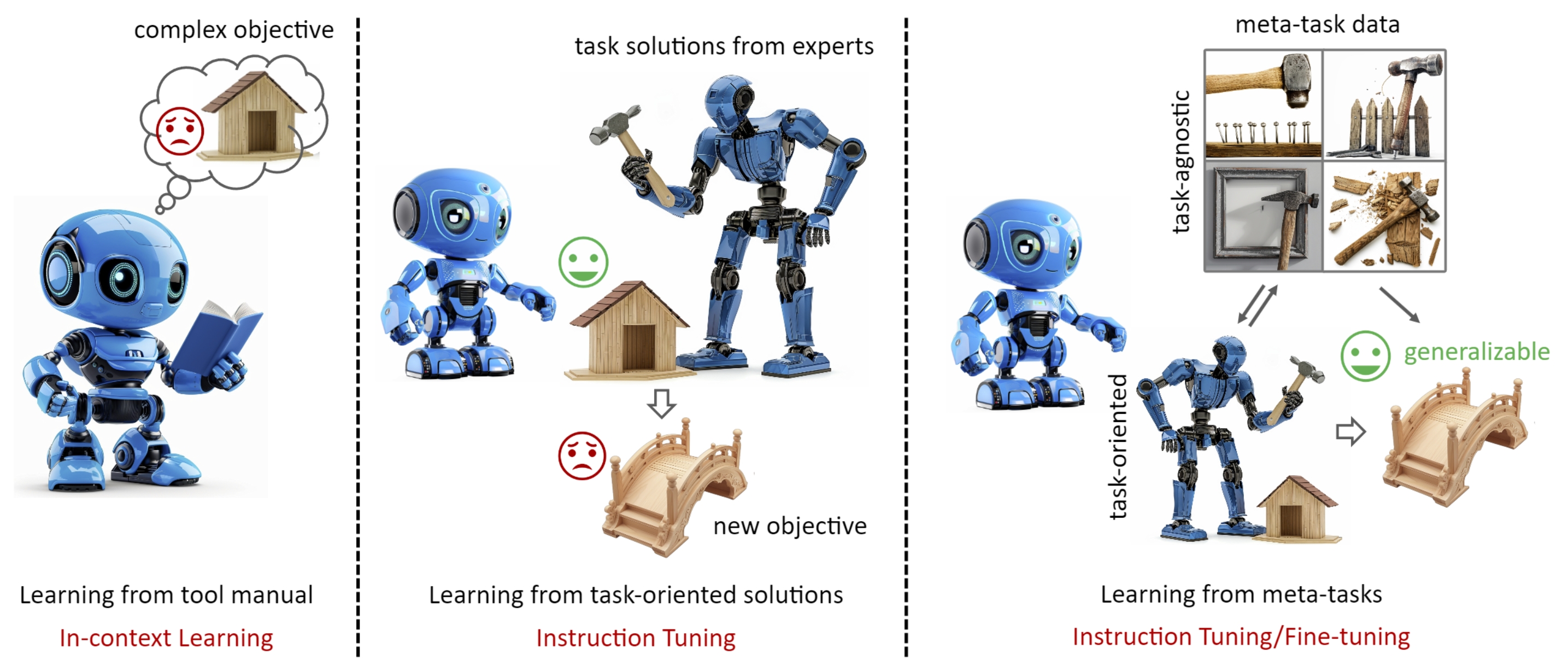}
    \caption{\textbf{Paradigm comparison} between existing tool learning methods and proposed meta-task augmentation. While the prevailing methods are limited in generalizing to complex scenarios or new tools, MetaTool enables gaining transferable tool understanding from task-agnostic knowledge.}
    \label{fig:enter-label}
\end{figure}

Empirically, human learners get familiar with tools such as hammers (e.g., for nailing and smashing) before engaging in complex construction. Generalizable tool use should be achieved based on the fundamental understanding of tools themselves that holds stable for different objectives, namely task-agnostic (illustrated in Figure 1). Naturally the formation of tool understanding can be disentangled from the learning of task solving. In this paper, we introduce \textit{MetaTool}, a general methodology that enables both complex tool mastery and unseen tool generalization on top of task-agnostic tool understanding. We design a set of meta-tasks inquiring about the causality of the toolset as an autonomous system and its functionality as a function. Given a callable toolset (e.g. APIs, functions), meta-task data is constructed in a scalable self-supervised way based on unsupervised or self-play tool executions. Augmenting task-oriented training with meta-task data, LLMs learn to solve problems while deepening the mastery of tools. We conduct experiments on both complex tool-oriented tasks and tool-augmented benchmarks, demonstrating that MetaTool significantly exceeds models trained merely on annotated solutions in both worlds and is competitive with the latest LLMs (ChatGPT) with the size of 8B. 
The overall contribution can be summarized in three-folds:

\begin{itemize}
    \item We introduce a new tool learning method that facilitates LLMs to master tools with task-agnostic tool understanding.
    \item We propose an integral set of self-supervised meta-tasks that dissect the tool execution process. Meta-tasks enable expert-free data generation and augmentation across tool-augmented and tool-oriented scenarios.
    \item Extensive evaluation on both tool-oriented tasks and tool-augmented benchmarks demonstrates the effectiveness and generality of MetaTool, narrowing the gap between open-source models and state-of-the-art LLMs.
\end{itemize}

\section{Approach}

In this section, we first formalize the task of using a close toolset and define 6 general meta-tasks that are key to tool understanding. Then we show how to construct datasets in an integral self-supervised way covering different scenarios. In the end, we describe several schemes to augment tool learning with meta-tasks.

\subsection{Self-supervised Meta-tasks for Tool Understanding}

\textbf{Problem formalization.} A tool-use task can be generally defined as a Markovian tuple $\left \langle \mathcal{S},\mathcal{A},\mathcal{T},g\right \rangle$, where $\mathcal{S},\mathcal{A}, \mathcal{T}$ is the state space, action space, and toolset, and $g$ is the goal state of the task. Toolset $\mathcal{T}=\{t\}_N$ consists of $N$ tools, each as a state transition function $s'=t(s,\theta)$ that formalizes the outcome of state change when feeding the input parameters $\theta$ into the tool. An action $a=\left \langle t,\theta \right \rangle \in \mathcal{A}$ specifies the tool and its input. As an autonomous agent, an LLM should iteratively respond with tool calling and inputs according to the state until it reaches the goal. Broadly, when the tools can not alter any external state, tool output like retrieval results can be regarded as the state, and the desired information is the goal $g$.

We enhance the tool understanding of the model with self-supervised surrogate (pretext) tasks instead of in-context descriptions or demonstrations. Formally, we regard tools as external systems that implement state transition mappings. Tool understanding, therefore, involves comprehending the perception-action process of these systems (referred to as tool execution) and should be generalizable to various task objectives. 

\begin{figure}[tb]
    \centering
    \includegraphics[width=1\linewidth]{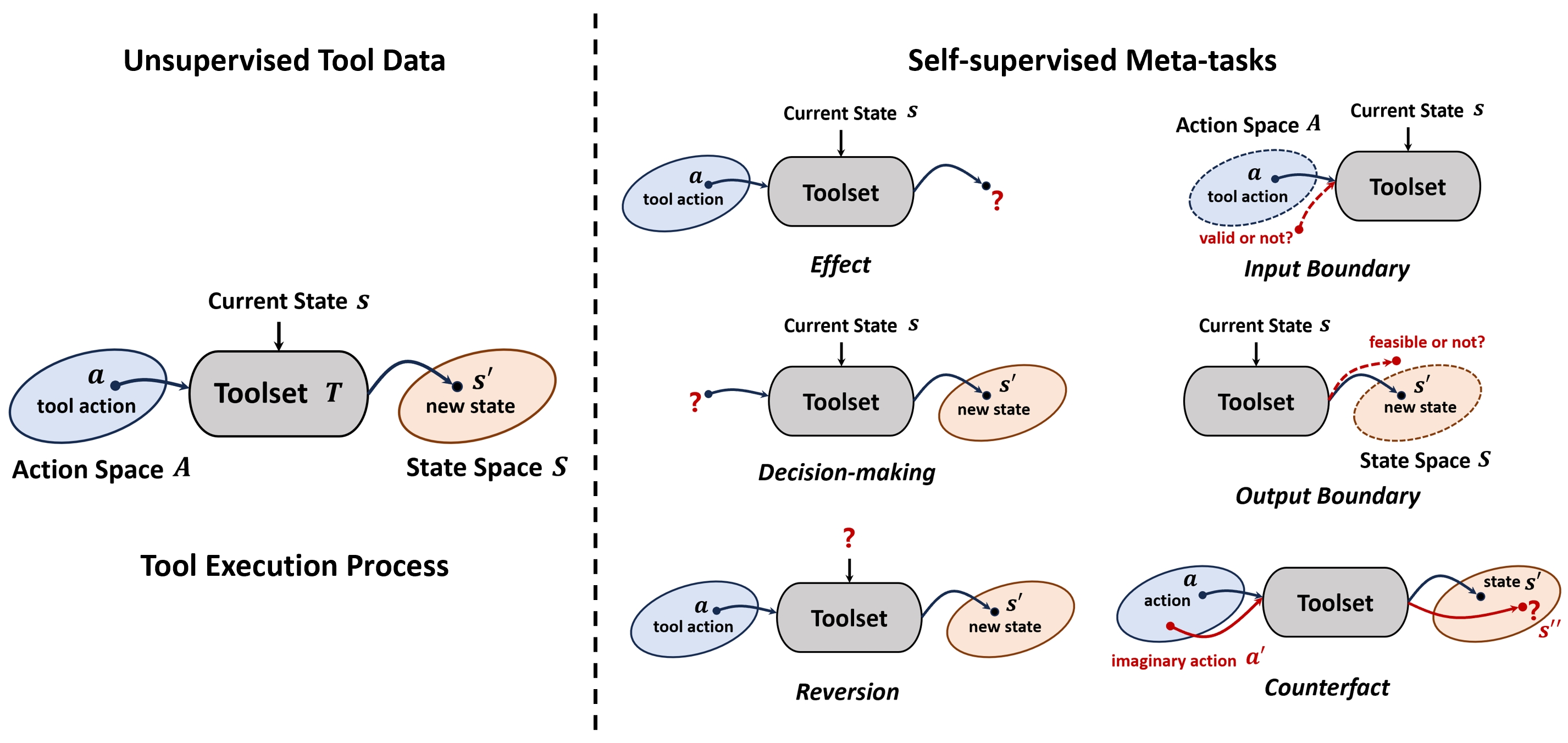}
    \caption{Illustration of developing self-supervised meta-tasks from unsupervised tool execution process.}
    \label{fig:enter-label}
\end{figure}

\textbf{Meta-task definition.} We begin with single-step tool execution $\mathcal{D}=\{s,a,s'\}$, peeling off the task goal $g$ context, which results as unsupervised data. Six surrogate tasks (meta-tasks) are designed based on the dataset $\mathcal{D}$. Basically, the model is required to predict masked elements of the execution process. It aligns with the idea of masked language models such as Bert \cite{devlin2018bert}, but in a broader and structural granularity. We define the meta-tasks as below (Figure 2):

\begin{itemize}
\item \textbf{Effect}: The model predicts the outcome state $P(s'|a,s)$ given the initial state and the action.
\item \textbf{Decision-making}: The model decides a feasible action $P(a|s,s')$ given the initial and outcome state. 
\item \textbf{Reversion}: The model deduces the initial state $P(s|a,s')$ given the action and the outcome state.
\item \textbf{Input Boundary}: The model determines whether an action can be successfully executed, namely whether the action falls in the valid action space, given the current state: $P(\mathbbm{1}_{s'\neq s}|a,s)$.
\item \textbf{Output Boundary}: The model determines whether a state can be reached with any action given the current state: $P(\mathbbm{1}_{\exists (t,\theta),s'=t(s,\theta)}|s,s')$.
\item \textbf{Counterfact}: The model predicts the new outcome state $P(s''|a,s',a')$ if a new action $a'$ were executed given that the current action $a$ results in the current outcome $s'$.
\end{itemize}

\textit{Effect}, \textit{decision-making}, \textit{reversion} meta-tasks emphasize the causality of a tool, regarding the action as the intervention to the state \cite{pearl2009causal,pearl2018book} and the outcome as the causal effect is determined by the tool mechanism. On top of that, \textit{counterfact} task is the composition of \textit{reversion} and \textit{effect}, further imagining the outcome altered from the fact in \textit{effect} task. This meta-task raises higher requirements on counterfactual reasoning \cite{bareinboim2015bandits,zhang2016markov}, an advanced form of causal reasoning that humans use to contemplate 'what if'. Moreover, tools implemented as APIs may receive non-executable inputs and result in ineffective outcomes. Thus the input and output domains are also unique features of a tool as a function. We consider \textit{input boundary} meta-task that emphasizes the tool affordance that refers to what actions can be executed considering the situation and the precondition. \textit{output boundary} meta-task emphasizes the functionality of tools, that is, what goals can and cannot be achieved given the current state.

\begin{figure}[t]
    \centering
    \includegraphics[width=1\linewidth]{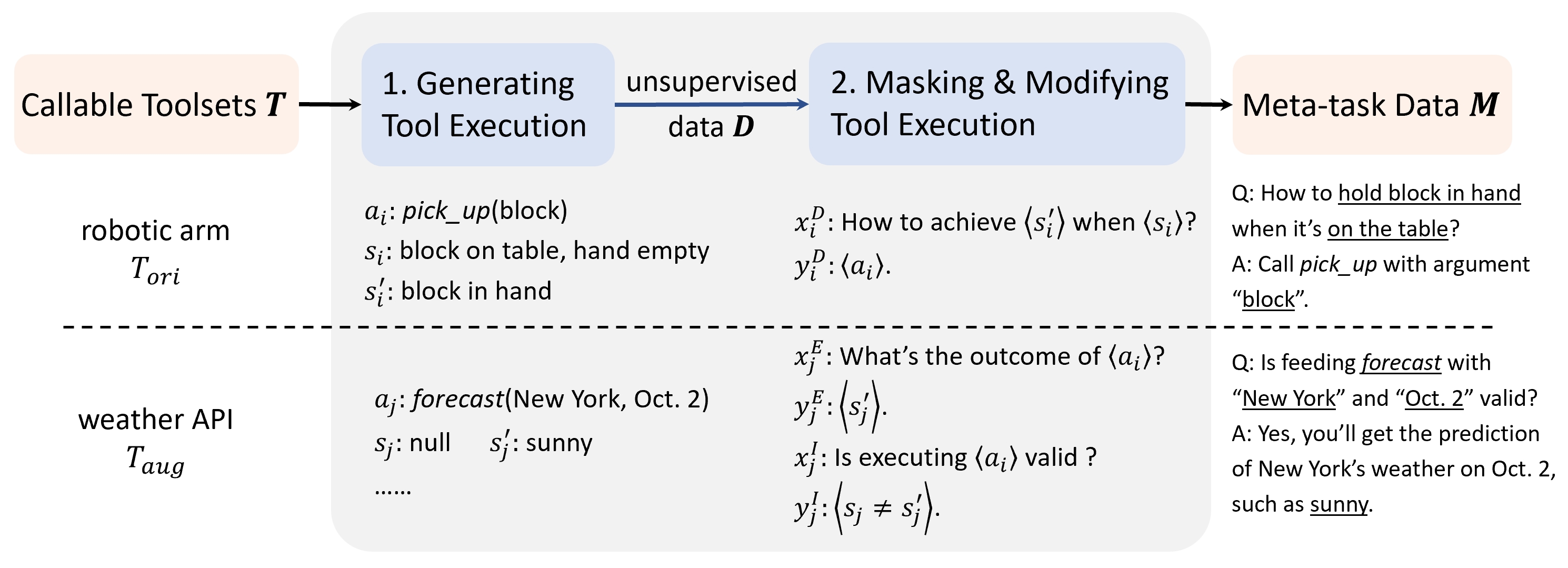}
    \caption{\textbf{Two-step approach to construct metaset.} We illustrate two examplified processes of both tool-oriented and tool-augmented scenarios, which don't require any expert annotation. $x_i^D, y_i^D$ denotes the $i$-th question-answer pair of \textit{decision-making} meta-task, et cetera.}
\end{figure}

\subsection{Metaset Construction} 

Based on the definitions, the dataset of meta-tasks (referred to as metaset) is generated as question-answering pairs to maintain the conversational skills of LLMs. To answer the meta-task questions, the trained model needs to understand the toolset mechanisms from the corresponding aspects. Given a set of reusable and callable tools $T$, the metaset $\mathcal{M}=\{x_n^m, y_n^m\}_{n=1}^N$ can be constructed in two steps, as illustrated in Figure 3, where $x_n^m, y_n^m$ is the $n$-th question-answer pair of meta-task $m$. First, we generate tool execution data $\mathcal{D}$ with the toolset. For a limited amount of tools and state space, stochastic sampling can be applied to initialize state $s \sim P(\mathcal{S})$ and action $a \sim P(\mathcal{A})$, and obtain the tool output $s'$. For large toolsets and diverse task scenarios that are hard to enumerate, we incorporate LLMs with self-play or tree search techniques to reduce redundant trials $a,s' \sim P(\mathrm{LLM}(s,g))$. Note that the LLM does not need to be proficient in tool tasks, as the execution data $\mathcal{D}$ is irrelevant to the task performance. Non-executable actions also contain valuable knowledge and can be transferred as invalid samples in \textit{input boundary} meta-task. The tool execution data should be sufficient to cover the various scenarios that may arise during the tasks. 

Second, for the $n$-th sample and meta-task $m$, we insert the variables of states and actions into $K$ sets of templates (diversified with GPT-4) to obtain diverse QA data.
\begin{equation}
    (x_n^m, y_n^m) = \mathrm{Mask}(a_n,s_n,s'_n,\Phi_k^m)
\end{equation}
, where $\Phi_k^m$ is the sampled template for meta-task $m$. Particularly, in tool-augmented scenarios, predicting the output of tools such as QA systems can be impractical. Nonetheless, the retrieval result itself is inherently meaningless (e.g., 'sunny'); however, it gains complete significance when combined with the context of tool calling (e.g., 'the weather in New York is sunny'), as showcased in Figure 3. Thus we modify the context into a more informative state in such scenarios by prompting LLMs $s^*_n=\mathrm{LLM}(s'_n,a_n,t)$, which is trivial for most language models. At last, we arrange multiple metaset pairs of the same toolset into multi-turn QA data as the metaset $\mathcal{M}$, in order to maintain multi-turn dialogue capacity.

\subsection{Tool Learning Augmented with Meta-tasks}

By incorporating meta-task data, we explore several approaches to augment the tool learning for the purpose of achieving task objectives: a) \textbf{In-context learning:} We randomly sample several demonstrations of each meta-task and add them to the system prompt to facilitate tool understanding in a training-free manner. Such task-agnostic knowledge includes the interpretation of rules, supplementing the solution demonstrations. b) \textbf{2-stage learning:} Since we aim to build the model's tool understanding as the foundation of tool-oriented learning, an intuitive idea is to train the LLM first on the metasets as the surrogate tasks and then on the solution data. In order to maintain the general ability of the model in the first stage, only the parameters of the query and value projection layers of the Transformer are updated instead of full-parameter training. c) \textbf{Data augmentation:} In this approach, we utilize the metaset as the augmented data of conventional instruction tuning methods that the metasets are mixed with solution data and the model is trained uniformly. The model trained on the mixed data is referred to as MetaTool.

\section{Experiments}

In this section, we evaluate our approach in both tool-oriented agent and tool-augmented chatbot scenarios. On the one hand, we fine-tune the LLM to master a specific toolset for achieving various complex objectives. On the other hand, we conduct large-scale instruction tuning to enable the model to generalize to new tasks and understand new tools through zero-shot documentation.

\subsection{Tool-oriented Scenarios}

\subsubsection{Task setup}

We adopt 3 tool-oriented tasks that emphasize complex tool execution and sequential planning. Among them SAW is newly designed while the other two are introduced from PlanBench \cite{valmeekam2024planbench}. The key challenge of these tasks is to understand the rules (preconditions) and the environmental dynamics caused by actions. The task definition and dataset construction are elaborated below.

\textbf{SpellAnyWord (SAW).} In this task, the agent needs to sequentially construct a string that contains the target string as a continuous substring. The initial state of the task is a void string. Two non-degradable tools (functions) are avaliable: \textbf{\textit{Add}}: to add two adjacent letters in the alphabet to the end of the current string. The tool input $\theta$ should be the preceding letter (e.g. passing 'a' to \textit{Add} on current string '' will result in 'ab'). \textbf{\textit{Swap}}: to swap the position of two adjacent letters in the current string. The input should be the preceding letter (e.g. passing 'a' to \textit{Swap} on 'ab' will result in 'ba'). An example task: The target string is 'any'. A successful action sequence can be [\textit{Add}('a'), \textit{Add}('n'), \textit{Add}('y'), \textit{Swap}('a'), \textit{Add}('o')], which will result in a state sequence ['ab', 'abno', 'abnoyz', 'banoyz', 'banyoz'] and the final string 'b\textbf{any}oz' has 'any' as a substring. To eliminate the basis from tokenization, we format each string as a list of letters in prompts throughout the task.

\textbf{BlocksWolrd (BW).} In this scenario, the agent needs to stack several blocks on the table into a target state with one hand. Only one block can be moved at a time. Two tools (functions) are avaliable: \textbf{\textit{Pick}}: to pick a block in the hand. The tool input should be the target block indicated by its color (e.g. \textit{Pick}('yellow')). Blocks cannot be picked if there are blocks on top of them or there's already a block in the hand. \textbf{\textit{Stack}}: to stack the block in the hand onto the target block or table. The input should be the color of the target block or 'table' (e.g. \textit{Stack}('white'), \textit{Stack}('table')). Blocks cannot be stacked on a block with another block already on top of it or there's no block in the hand.

\textbf{Logistics (LOG).} The agent needs to solve a logistics problem by arranging trucks and airplanes to transport the package to the target location. Locations are grouped by cities. Trucks can be used to move packages between locations in the same city and planes can be used to move packages between cities. Two tools (functions) are available: \textbf{\textit{Truck}}: to transport the truck and the package (if there is any) from one location to another.  \textbf{\textit{Plane}}: to transport the airplane and the package (if there is any) from one location to another. The tool input should be the starting and ending location indicated by numbers. (e.g. \textit{Truck}(1,2), \textit{Plane}(2,4)). An action is invalid when there is no truck or airplane at the starting location.

\textbf{Datasets collection.} For the SAW task, we randomly sample 2k target strings (from 2 letters to 10 letters) as task goals. We modify the BlocksWorld and Logistics tasks from PlanBench into the tool-use version, thus 2k goals for each task are adopted following the original configuration. Optimal action sequences are obtained with heuristic strategy as the solution data. For each annotated action, we generate a thought with GPT-generated templates. The thoughts analyze the situation and what to do following ReACT \cite{yao2022react} to leverage the model's reasoning ability. Thus the solution contains a sequence of thought-tool-input tuples.

\subsubsection{Implementation details}

Our model is fine-tuned based on LLaMA3-8b-instruct \cite{llama3modelcard} with parameter-efficient fine-tuning method Qlora \cite{dettmers2024qlora} on 8 A100 GPUs. We utilize the instruction tuning version of LLaMA3 since comprehending tool-oriented tasks with specific objectives is the basis of tool understanding and use. For each task, we train the model on 10k meta-task data and 10k solution data for 3 epochs with AdamW optimizer and the learning rate of 2e-4. The models are tested in a simulated environment that receives the action of using a tool and returns the outcome and current state. We evaluated the model performance on 100 unseen cases of each three tasks. 

\begin{table}[tb]
\centering
    \begin{minipage}{0.44\textwidth}
        \centering
        \tabcolsep=4pt
        \begin{tabular}{l|c|c|c}
        \toprule
            \textbf{Models} & \textbf{SAW} & \textbf{BW} & \textbf{LOG} \\
        \midrule
             ChatGPT & 22.6& 23.3& 50.4\\
             ChatGPT-ICL &20.2 &20.5&43.6\\
             GPT-4 & 28.6 & 43.0& 46.6\\
             GPT-4-ICL & 27.4&40.0 &37.0 \\
        \midrule
             Vicuna-7b &4.8&5.5&0.0 \\
             LLaMA3-8b-instruct & 6.0& 6.7& 6.0\\
             LLaMA3-solution &9.5&19.2&8.2\\
             LLaMA3-ICL & 4.8&17.8&2.0 \\
             LLaMA3-2-stage &9.5&21.9&12.3\\
             MetaTool &32.1&37.5& 30.1\\
        \bottomrule
        \end{tabular}
        \caption{\textbf{Results on tool-oriented tasks.} ICL: in-context learning with meta-task demonstrations. ChatGPT and GPT-4 are provided with tool documentation and few-shot demonstrations.}
    \end{minipage}
    \hfill
    \begin{minipage}{0.53\textwidth}
        \centering
        \tabcolsep=4pt
        \begin{tabular}{ccccccc|c|c|c}
        \toprule
            E & D & R & I & O & C & S & \textbf{SAW} & \textbf{BW} & \textbf{LOG} \\
        \midrule
              \usym{1F5F4}&&&&& &  &9.5&27.0&11.0 \\
              &\usym{1F5F4}&&&& &  &18.5&29.0&9.0 \\
              &&\usym{1F5F4}&&& &  &27.3&35.0&21.0 \\
              &&&\usym{1F5F4}&& &  &17.3&32.0&18.0 \\
              &&&&\usym{1F5F4}& &  &16.1&32.0&6.0 \\
              &&&&&\usym{1F5F4} &  &19.6&37.0&14.0 \\
              &&&&&&\usym{1F5F4}   &15.5&31.0& 10.0\\
        \midrule
             \usym{1F5F8}&\usym{1F5F8}&\usym{1F5F8}&\usym{1F5F8}&\usym{1F5F8}&\usym{1F5F8} & \usym{1F5F8} &32.1&37.5& 30.1\\
        \bottomrule
        \end{tabular}
        \caption{\textbf{Ablation on tool-oriented tasks.} E: effect meta-set, D: decision-making meta-set, R: Reversion meta-set, I: input boundary meta-set, O: output boundary meta-set, C: counterfact meta-set, S: solution dataset. The crossings denote removing the training data of the corresponding meta-tasks.}
    \end{minipage}
\end{table}
\vspace{-0.5cm}

\subsubsection{Results analysis}

\textbf{Overall comparison.} We evaluate the success rate (SR\%) of completing each task and show the performances of several models in Table 1. Overall, SOTA closed-source LLMs show impressive zero-shot performance on tool-oriented tasks compared with open-source LLMs including LLaMA3 and Vicuna. By training on both meta-tasks and solution data, our model MetaTool gains significant improvement (+20.9\%SR on average) compared with mere training on solution data (LLaMA3-solution). MetaTool also surpasses GPT-4/ChatGPT in the SAW/BW tasks (+3.5\%/14.2\%SR). ChatGPT represents the model of GPT-3.5-turbo throughout our experiments. Both GPT and LLaMA3 show weaker performances when provided with meta-task demonstrations (ICL) since demonstrating limited cases can be redundant or misleading without proper design. LLaMA3-2-stage that trained on meta-tasks first gains limited improvement compared with the baseline. We conjecture that learning meta-tasks without practicing tool use (training on action sequences) cannot effectively facilitate tool-use ability with tool understanding. Also fine-tuning with specific QA data may affect the basic linguistic ability of the model. The overall results show that LLMs (including the most powerful ones like GPT-4) still have difficulties conquering complex tool using tasks, especially in planning with tools.

\textbf{Ablation study.} We study the ablation of different data components and report the performances in Table 2. It's worth noticing that merely training on meta-tasks can improve the model's zero-shot performance on tool-oriented tasks (line 7), contrary to providing demonstrations of meta-tasks in the system prompt (LLaMA3-ICL in Table 1). When removing QA data from each meta-task, the model performance shows varying degrees of degradation, which verifies the profits of meta-tasks. The meta-tasks of \textit{effect} and \textit{decision-making} have a relatively greater influence on the model's tool understanding capability. Theoretically, these meta-tasks emphasize the causal mechanism of tools that is more fundamental than others.

\subsection{Tool-augmented Scenarios}

\subsubsection{Task setup}

Among the various tool/function calling benchmarks, we study our method on two of the most influential ones: ToolBench \cite{qin2023toolllm} and Berkeley Function Calling Leaderboard (BFCL) \cite{berkeley-function-calling-leaderboard}. \textbf{ToolBench} contains diverse user requests with a massive amount (over 16k) of real-world API tools, which are publicly available on the RapidAPI website. The testset is categorized into six distinct groups and contains 1200 instructions (200 each): I1-inst., I1-tool, I1-cat., I2-inst, I2-cat., and I3-inst. Groups labeled with I1, I2, I3 include single-tool tasks, intra-category multi-tool tasks, and extra-category multi-tool tasks, respectively. Groups labeled with “inst.”, "tool", "cat." include unseen user instructions, unseen tools, and unseen categories (e.g. sports, entertainment) of tools, respectively. Two evaluation metrics are designed based on ChatGPT: (1) Pass Rate, calculated by the proportion of instructions successfully completed within a limited budget; (2) Win Rate measured by asking a ChatGPT evaluator to select its preference for two solution paths. For each user instruction (e.g. ``Can you recommend some popular restaurants within 5km to hold a party?''), the model calls an API and responds to the query based on the tool output.

ToolBench also provides 126k instruction-solution pairs for training, which are generated with GPT4 and depth-first tree search (DFS). On top of that, we extract unsupervised tool execution data and generate 650k self-supervised data of meta-tasks following the procedure in Figure 3. We then conduct instruction tuning based on the mixed data and LLaMA3-8B-instruct model, trained for 1 epoch to avoid overfitting. Both the solution data and the meta-task data share the same loss setting as we construct the metaset as QA pairs. The context-aware states are generated with the open-sourced LLaMA3-70B-instruct model. All evaluated LLMs are prompted in the ReACT manner to leverage their reasoning ability. Other model implementation details are in line with that described in section 4.1.2.

\textbf{BFCL benchmark} is established mainly for the purpose of zero-shot evaluation without holistic training data. The benchmark contains 4251 testing cases in total and is categorized into non-live (self-designed), live (user-contributed), multi-turn, and Hallucination (relevance or irrelevance determination) groups. The model performance is measured by action accuracy with Abstract Syntax Tree (AST) \cite{patil2023gorilla}.
We test the MetaTool model trained with ToolBench data on BFCL to evaluate the zero-shot ICL ability and generality of our methodology. 

\begin{table}[tb]
\centering
\tabcolsep=3.6pt
\footnotesize
    \begin{tabular}{l|cc|cc|cc|cc|cc|cc|cc}
    \toprule
        \multirow{2}{*}{\textbf{Models}} & \multicolumn{2}{|c|}{\textbf{\underline{I1-Inst.}}} & \multicolumn{2}{c|}{\textbf{\underline{I1-Tool}}} & \multicolumn{2}{c|}{\textbf{\underline{I1-Cat.}}} & \multicolumn{2}{c|}{\textbf{\underline{I2-Inst.}}} & \multicolumn{2}{c|}{\textbf{\underline{I2-Cat.}}} & \multicolumn{2}{c|}{\textbf{\underline{I3-Inst.}}} & \multicolumn{2}{c}{\textbf{\underline{Averages}}} \\
        & Pass & Win & Pass & Win & Pass & Win & Pass & Win & Pass & Win & Pass & Win & Pass & Win \\
    \midrule
         ChatGPT & 41.5&-&41.0&-&41.0&-&34.5&-&46.5&-&22.0&-&37.8&- \\
         Claude-2 & 5.5&31.0&3.5&27.8&5.5&33.8&6.0&35.0&6.0&31.5&14.0&47.5&6.8&34.4 \\
         GPT-4 & \textbf{53.5}&\textbf{60.0}&\textbf{50.0}&\textbf{58.8}&\textbf{53.5}&\textbf{63.5}&\textbf{67.0}&\textbf{65.8}&\textbf{72.0}&\textbf{60.3}&\textbf{47.0}&\textbf{78.0}&\textbf{57.2}&\textbf{64.4}\\
         ToolLLaMA-2 & 25.0&45.0&29.0&42.0&33.0&\underline{47.5}&30.5&50.8&31.5&41.8&25.0&55.0&29.0&47.0 \\
    \midrule
         LLaMA3-8B-inst. & 0.0&0.0&0.0&0.0&0.1&0.0&0.0&0.0&0.1&0.1&0.0&0.0&0.0&0.0 \\
         LLaMA3-solution & 32.1&45.3&39.0&43.9&36.4&43.0&40.1&52.5&40.1&43.4&35.6&61.8&37.2& 48.3\\
         MetaTool (8B) & \underline{42.5}&\underline{52.1}&\underline{41.8}&\underline{51.3}&\underline{43.3}&46.1&\underline{52.0}&\underline{54.9}&\underline{50.0}&\underline{54.0}&\underline{45.5}&\underline{74.5}&\underline{45.9}&\underline{55.5}\\
    \bottomrule
    \end{tabular}
\caption{\textbf{ToolBench results.} ChaGPT doesn't have the Win Rate score since all other Win Rates are measured by comparing with its solution paths.}
\end{table}

\begin{table}[t]
\centering
\tabcolsep=1pt
\footnotesize
    \begin{tabular}{l|cccc|cccc|c|cc|c}
    \toprule
        \multirow{2}{*}{\textbf{Models}} & \multicolumn{4}{|c|}{\textbf{\underline{Non-live}}} & \multicolumn{4}{c|}{\textbf{\underline{Live}}} & \textbf{\underline{Multi Turn}} & \multicolumn{2}{c|}{\textbf{\underline{Hal.}}} & \textbf{\underline{Simple}}\\
        & simple & multiple & parallel & M\&P & simple & multiple & parallel & M\&P & base & rel. & irrel. & \textbf{\underline{Ave.}}\\
    \midrule
         GPT-4-turbo & 60.6&91.0&90.0&89.0&67.8&74.5&75.0&62.5&33.5&70.7&79.8&54.0 \\
         o1-mini & 68.9&89.0&73.5&70.5&62.8&65.1&68.8&58.3&16.0&46.3&88.7&49.2 \\
         Hermes-2 (8B) & 61.3&82.5&75.5&75.0&55.8&53.1&43.8&41.7&1.5&51.2&62.7&39.5\\
    \midrule
         LLaMA3-solution &71.3 &64.0&13.5&10.0&56.6&34.9&37.5&12.5&5.0&100.0&8.3&44.3 \\
         MetaTool (8B) & 78.3&55.0&66.0&63.5&58.1&50.1&18.8&37.5&6.5&100.0&25.4&47.6\\
    \bottomrule
    \end{tabular}
\caption{\textbf{BFCL results.} M\&P denotes the test set of multiple parallel. Hal., rel., and irrel. represent the relevance and irrelevance set of the hallucination group. Simple Ave. denotes the average accuracy of non-live simple, live simple, and multi-turn for fair comparison.}
\end{table}


\subsubsection{Results analysis}

As the ToolBench results shown in Table 3, MetaTool (8B) achieves the second-best performance across all groups merely behind GPT-4, superior to other models including ChatGPT (+8.1\% pass rate) and training-based ToolLLaMA-2 (7B) \cite{qin2023toolllm} (+16.9\% pass rate, +8.5\% win rate). Especially, our method significantly improves the performance of LLaMA3 (originally incapable) compared to LLaMA3-solution which is merely trained on solution data (+8.7\% pass rate, +7.2\% win rate). The comprehensive advantages show the effectiveness of meta-task augmentation. Besides, the notable superiority of GPT-4 can be partly attributed to the fact that all the training and testing instructions are generated with itself. Thus GPT-4 may be more familiar with the distribution and inner motivation of these instructions.

Table 4 shows the zero-shot performance on the BFCL benchmark. It's worth noticing that except for the sets of non-live simple, live simple, and multi-turn base, zero-shot comparison on other test sets is less fair, since MetaTool is merely trained in ToolBench scenarios with a unique task pattern (fixed system prompt) that the model calls a single tool once a time then waits for the tool output. For example, in the "multiple" tasks LLMs are asked to call multiple tools in one response. Nonetheless, we modify the parser of MetaTool to continue generating tokens to fit the requirements of multiple and parallel scenarios. In the "irrelevance" tasks, LLMs have access to tools irrelevant to the instruction and should give up calling any tools, which never occurs in ToolBench scenarios. Therefore in the first place, we count the average accuracy (Simple Ave. in Table 4) of three fair sets and witness that MetaTool surpasses LLaMA3-solution (+3.3\%) and Hermes-2-theta \cite{Hermes-2-Theta-Llama-3-8B} (+8.1\%) which is also trained based on LLaMA3-8B-instruct, and is close to the latest OpenAI o1-mini (-1.6\%). It's also impressive that MetaTool obtains the highest 78.3\% accuracy on the non-live simple set, 17.7\% higher than the 1st rank model GPT-4-turbo. Despite the transferring barriers, MetaTool still achieves moderate performance on the other sets (e.g. multiple, parallel, hallucination) with an average of 52.0\% accuracy, significantly higher than the 35.1\% of LLaMA3-solution (+16.9\%). Overall, the zero-shot results on BFCL clearly demonstrate the exceptional generalizability of MetaTool.  With additional data augmentation from diverse scenarios, MetaTool has the potential for significant improvement.

\vspace{-0.3cm}

\begin{figure}[tb]
    \centering
    \includegraphics[width=0.9\linewidth]{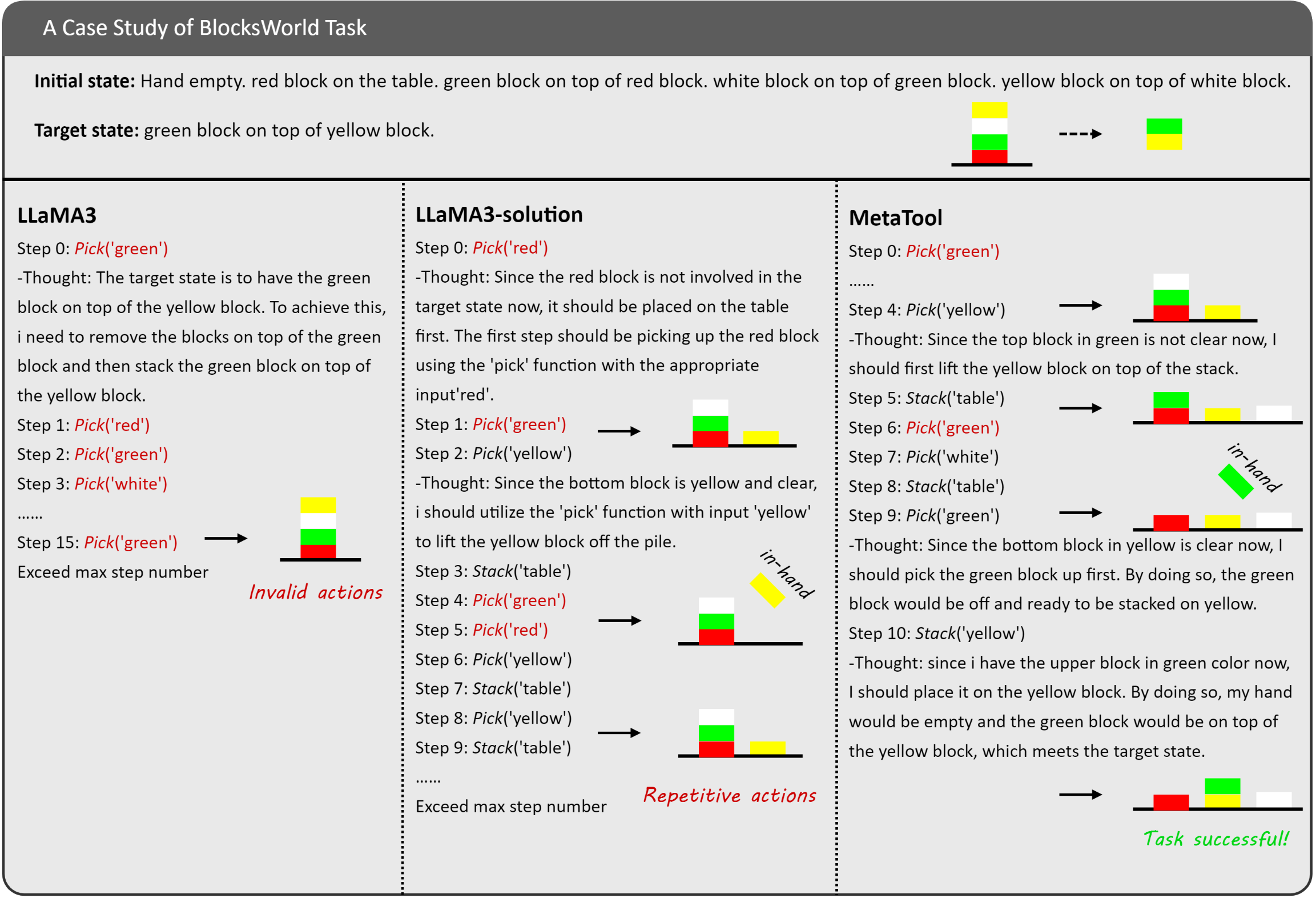}
    \caption{Case study of MetaTool compared with 2 baselines on BlocksWorld task. Actions in red denote invalid ones (e.g. pick up a block at the bottom). LLaMA3-solution is the LLaMA model trained on task solution data.}
    \label{fig:enter-label}
    \vspace{-0.3cm}
\end{figure}

\subsection{Qualitative Case Study}

As showcased in Figure 4, the agent is required to construct stacks containing a green block on top of a yellow block from a pile of 4 blocks. With mere descriptions of tools in the prompts, LLaMA3 fails to understand the precondition of using tools resulting in invalid actions. Training on tool-oriented solution data, LLaMA3-solution attempts to lift the yellow block successfully but fails to sequentially achieve the task goal and falls into repetitive loops. The proposed MetaTool model achieves the target state with an effective action sequence (although still not the optimal efficiency) and corresponding reasoning. These 3 models correspond to the 3 paradigms illustrated in Figure 1. The results show that LLMs can learn tool use better on the basis of robust tool understanding.

\section{Related Works}

\subsection{Tool learning}

Recent studies have shed light on the potential of utilizing tools to augment LLMs with external factual knowledge \cite{qin2023webcpm,nakano2021webgpt,song2023restgpt,hao2024toolkengpt,shen2024hugginggpt,gao2023assistgpt,wu2023visual,qian2023creator,zhuang2024toolqa,schick2024toolformer} which is targeted at tool-augmented question-answering scenarios, towards the `tools for AI' purpose in general. On the other side, With the burgeoning intelligence in reasoning and perception, LLMs' tool-use capability can be widely applied in the automation of various domains including Embodied AI \cite{wang2024camp,wang2024interactive}, web manipulation \cite{rawles2023android,hong2024cogagent,yang2023appagent,deng2024mind2web,he2024webvoyager,zhou2023webarena}, and image/video editing \cite{wang2024lave,argaw2022anatomy,hang2024cca,fu2023guiding}.This line of work is intended for tool-oriented planning scenarios for the `AI for tools' purpose. Effectively mastering complex tools challenges the model to comprehend the precondition and potential outcome of using tools. In this paper, we aim to facilitate LLMs for both tool-oriented and tool-augmented tasks by learning robust tool understanding.

\subsection{Tool understanding}

As noted by \cite{hernik2009functional}, when learning to utilize a specific tool, children perceive it as an object with particular functions, engaging in a cognitive process to understand its purpose and operation. Analogously, a comprehensive understanding of the tools’ functionalities is indispensable for enabling the controller to use tools proficiently. In real-world scenarios, tools are typically accompanied by a manual (or tutorial), which provides sufficient
relevant details about their functionalities and usage. Endowed with strong few-shot learning \cite{brown2020language} and zero-shot learning \cite{wei2021finetuned} capabilities, foundation models can be prompted to unravel tools’ functionalities and comprehend how to use them. To this end, we can construct suitable task-specific
prompts either through manual design \cite{vemprala2024chatgpt} or retrieval \cite{zhou2022docprompting}. However, prompting is restricted by input context length, thus the situation may be more challenging with multiple complex tools with long descriptions. While most training-based tool learning methods rely on extensive expert-annotated solution data for goal-oriented tasks, the knowledge contained in the tool execution process itself remains unutilized. We propose a self-supervised data augmentation method to efficiently endow LLMs the comprehension of a set of tools.

\vspace{-0.1cm}
\section{Conclusion}
\vspace{-0.3cm}

In this work, we introduced MetaTool, an efficient and generalizable method that facilitates tool learning with task-agnostic comprehension. This is achieved by deriving self-supervised meta-task data from tool execution actions. Augmented the meta-tasks into either complex toolset fine-tuning or large-scale instruction tuning, our model exhibits sophisticated tool mastery as well as generality in in-context learning. Evaluated on multiple tool use benchmarks, MetaTool outperforms models trained on expert solutions and showcases comparable performance with ChatGPT in a size of 8B.

\bibliography{iclr2025_conference}
\bibliographystyle{iclr2025_conference}


\end{document}

%% file: math_commands.tex
\usepackage{amsmath,amsfonts,bm}

\def\eqref#1{equation~\ref{#1}}

\def\1{\bm{1}}

\DeclareMathAlphabet{\mathsfit}{\encodingdefault}{\sfdefault}{m}{sl}
\SetMathAlphabet{\mathsfit}{bold}{\encodingdefault}{\sfdefault}{bx}{n}

